# A Review of Methodologies for Natural-Language-Facilitated Human-Robot Cooperation

Rui Liu, Xiaoli Zhang*

*Abstract-* **Natural-language-facilitated human-robot cooperation (NLC) refers to using natural language (NL) to facilitate interactive information sharing and task executions with a common goal constraint between robots and humans. Recently, NLC research has received increasing attention. Typical NLC scenarios include robotic daily assistance, robotic health caregiving, intelligent manufacturing, autonomous navigation, and robot social accompany. However, a thorough review, that can reveal latest methodologies to use NL to facilitate human-robot cooperation, is missing. In this review, a comprehensive summary about methodologies for NLC is presented. NLC research includes three main research focuses: NL instruction understanding, NL-based execution plan generation, and knowledge-world mapping. In-depth analyses on theoretical methods, applications, and model advantages and disadvantages are made. Based on our paper review and perspective, potential research directions of NLC are summarized.**

*Index Terms-* **natural language, human-robot cooperation, NL instruction understanding, NL-based execution plan generation, knowledge-world mapping**

## I. INTRODUCTION

### A. Background

Attracted by the naturalness of natural language (NL) communications among humans, intelligent robots start to understand NL to develop intuitive human-robot cooperation in various task execution applications [1][2]. Natural-language-facilitated human-robot cooperation (NLC) has received increasing attention in human-involved intelligent robotics research over the recent decade. By using NL, human intelligence at high-level task planning and robot physical capability—such as force [3], precision [4], and speed [2]—at low-level task executions are combined to perform intuitive cooperation [5][6].

Due to the advantages of naturalness, information richness and standardized linguistic structures, NLC has been widely explored in areas, including daily assistance [5], medical caregiving [6][7], manufacturing [8], indoor or outdoor navigation [9], and social accompany [10]. Typical areas using NLC systems are shown in Fig. 1.

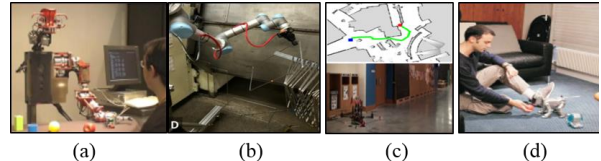

(a) (b) (c) (d)

Fig. 1. Promising areas using natural-language-based human-robot cooperation. (a) is daily robotic assistance using NL [11]. A robot categorized daily objects with human NL instructions. (b) is autonomous manufacturing using NL [12]. An industrial robot welded parts with human's oral requests. (c) is robotic navigation using NL [13]. A quadcopter navigated in indoor environments with human's oral guidance. (d) is social accompany [14]. A pet dog is playing balls with a human with socialized verbal communications.

### B. Motivation and systematic overview

Advancements of NLP support an accurate understanding of the task in NLC. Advancement of a robot's physical capability support increasingly improved task execution in NLC. With supporting technique from both NLP and robot execution, NLC has been developed from low-cognition-level symbol matching control, such as using "yes/no" to control robotic arms, to high-cognition-level task understanding, such as identifying a plan from the description "go straight and turn left at the second cross."

NLC research is regularly published in international journals, such as IJRR [15], TRO [16], AI [17] and KBS [18], and international conferences such as ICRA [19], IROS [20] and AAAI [21]. By using keywords 'NLP, human, robot, cooperation, speech, dialog, natural language', about 1200 papers were retrieved from Google Scholar [22], then with a focus of NL-facilitated human-robot cooperation, about 420 papers were finally kept. The publication trend is shown in Fig. 2, where the increasing significance of NLC is reflected by steadily increasing publication numbers.

Compared with existing review papers about human-robot cooperation using communication manners such as gesture and pose [23][24], action and motion [25], and tactile [26], a review paper about NLC, which is human-robot cooperation using NL communication manner, is lacking. Given the huge potentials and increasingly received attention, it is necessary to make a summary of state-of-the-art NLC methodologies in wide-range

______________________________
Rui Liu and Xiaoli Zhang are with the Intelligent Robotics and Systems Lab (IRSL), Colorado School of Mines, Golden, CO 80401 USA. (e-mail:rliu@mines.edu, xlzhang@mines.edu.).
*Corresponding author.

domains, revealing current research progress and signposting future NLC research. The organization of this paper is shown in Fig. 3.

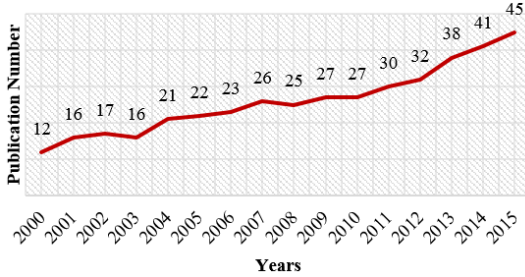

Fig. 2. The annual amount of NLC-related publications since the year 2000 according to our paper review. In the past 15 years, the number of NLC publications are steadily increasing and reaching a history-high level in current time, revealing that NLC research is encouraged by other researches such as robotics and NLP.

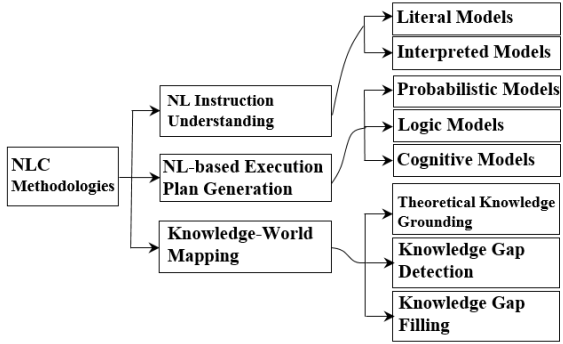

Fig.3. Organization of this review paper. This review systematically summarized methodologies for using NL to facilitate human-robot cooperation. Three main researches are introduced as NL instruction understanding, NL-based execution plan generation, and knowledge-world mapping. In each research, typical models, application scenarios, model comparison and open problems are summarized.

## II. FRAMEWORK OF NLC REALIZATION

Realization of NLC is challenging due to the following aspects. First, human NL is abstract and ambiguous. It is hard to understand humans accurately during task assignments, impeding natural communications between a robot and a human. Second, NL-instructed plans are implicit. It is difficult to reason appropriate execution plans from human NL instructions for effective human-robot cooperation. Third, NL-instructed knowledge is information-incomplete and real-world inconsistent. It is difficult to map sufficient theoretical knowledge into the real-world for supporting successful NLC. To solve these problems for effective and natural NLC, mainly three types of researches have been done. First, to accurate understand assignments during NLC, *NL instruction understanding* research has been done to build semantic models for extracting cooperation-related knowledge from human NL instructions. Second, to reason a robot's execution plans from human NL instructions, *NL-based execution plan generation* research has been done to create various reasoning mechanisms for identifying human requests and formulate robot execution strategies. Third, to map NL-instructed theoretical knowledge to real-world situations for practical cooperation, *knowledge-world mapping* research has been done to recommend the missing knowledge and correct the real-world inconsistent knowledge for realizing NLC in various real-world environment.

## III. NL INSTRUCTION UNDERSTANDING

NL instruction understanding enables a robot to receive human-assigned tasks, identify human-preferred execution procedures, and understand the surrounding environment from abstract and ambiguous human NL instructions during NLC. By improving the robot's understanding towards the human, the accuracy and naturalness during NLC are improved. To intuitively understand human NL expressions with an environment awareness, two types of semantic analysis models were developed: literal models and interpreted models. For both literal models and interpreted models, cooperation related information is explicitly or implicitly extracted indicated by humans. The difference between them, however, is the information source. The literal models only extract information from human NL instructions; while the interpreted models will also extract information from human's surrounding environment. With literal models, the robot understands tasks merely by following human NL instructions; while with interpreted models, robots understand tasks by critically thinking about cooperation-related practical environment conditions, becoming situation aware.

*A. Models*

From the model construction perspective, to analyze meanings of human NL instructions in NLC, literal models mainly use literal linguistic features, such as words, Part-of-Speech (PoS) tags, word dependencies, word references and sentence syntax structures, shown in Fig. 4; interpreted models mainly use interpreted linguistic features, such as temporal and spatial relations, object categories, object physical properties, object functional roles, action usages, and task execution methods, shown in Fig. 5. Literal linguistic features were directly extracted from human NL instructions, while interpreted linguistic features were indirectly inferred from commonsense based on NL expressions.

(i). Literal models

With regards to involvement manners of literal linguistic features, literal models are categorized into the following types. (1) Grammar model. Literal linguistic feature patterns such as "action+destination" are manually defined. (2) Association model. Literal linguistic features are mutually associated with commonsense knowledge.

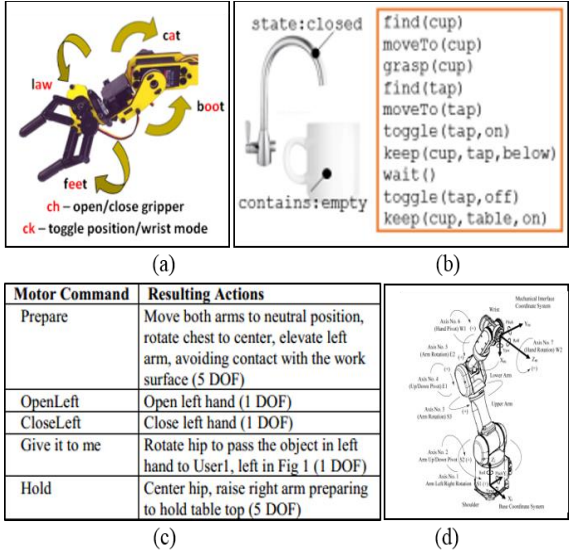

Fig.4. Typical literal models for NL instruction understanding. (a) [27], and (b) [29] are grammar models. In (a), the robotic arm's motion was controlled by predefined vowels in human speech. In (b), object manipulation methods are defined as linguistic structures such as "find(cup), grasp(milkbox), …". (c) [34] and (d) [33] are association models. NL expression such as "OpenLeft" was interpreted as specific parameter "open left hand for 1 DOF" for robotic arms.

To initially identify key cooperation-related information, such as goal, tool usage, and action sequences, from human NL instructions, grammar patterns are defined to build *grammar models*. Grammar patterns refer to keyword combinations, PoS tag combinations and keyword-PoS tag combinations [30][31]. By using these grammar models, robot behaviors will be triggered by the grammars mentioned in human NL instructions. Some grammar patterns explored execution logics. For example, verbs and nouns were combined to describe a type of actions such as V(go) + NN(Hallway) and V(grasp) + NN(cup) [32][35][36]. Some grammar patterns explored temporal relations, such as the if-then relation "if door open, then turn right" and the step1-step2 relation "go -- grasp" [37][38]. Some grammar patterns explored spatial relations, such as the IN relation "cup IN room" and the CloseTo relation "cup CloseTo plate" [9][39]. The rationale of the grammar model is that sentences with a similar meaning have similar syntax structures. Similarity of NL meanings was calculated by evaluating the syntax structure similarity.

To understand abstract and implicit NL execution commands during cooperation, *association models* were developed by associating different literal linguistic features together to extract new semantic meanings. Essentially, the association model exploited existed knowledge by creating high-level abstract knowledge from low-level detailed knowledge. One typical association model is a probabilistic association model. Informative literal linguistic features in NL instructions were correlated with other informative keywords by using probability likelihoods computed from human communications. Typical works are as follows. (1). Learning from previous human execution experiences. Cooperation-needed actions are inferred based on mentioned tasks, locations, and their probabilistic associations [40]. (2) Learning from daily commonsense. Quantitative dynamic spatial relations such as "away from, between, …" have been associated with its corresponding NL expressions based on their probabilistic relations [41]; general terms such as 'beverage' are specified to 'juice' according to cooperation types and task-object probabilistic relations [42]. With this probabilistic association model, the uncertainty in NL expressions was modeled, disambiguating NL instructions and improving a robot's adaptation towards different human users with various NL expressions. Another typical association model is an empirical association model. High-level abstract literal linguistic features, such as ambiguous words and uncertain NL phrases, are empirically specified by low-level detailed literal features such as action usage, sensor values, and tool usages. The rationale is that general knowledge could be recommended for disambiguating ambiguous NL instructions in specific situations. Compared with probabilistic association models, which use objective probabilistic calculation, empirical association models use subjective empirical association. Typical usages include the following types. (1). By defining sensor value ranges as ambiguous NL descriptions, such as "slowly, frequently, heavy", ambiguous execution-related NL expressions were quantitatively interpreted, making ambiguous NL expressions sensor-perceivable [38][43]. (2). By integrating key aspects, such as execution preconditions, action sequences, human preferences, tool usages, and location information, into abstract NL expressions—such as 'drill a hole'—human instructed high-level plans were specified into detailed robot-executable plans—such as 'clean the surface,' or 'install a screw' [4][9][31][36][44]. (3). By using discrete fuzzy statuses—such as 'close, far, cold, warm'—to divide continuous sensor data ranges, unlimited objective sensor values were 'translated' into limited subjective human feelings, such as "close to the robot, day is hot", supporting a human-centered task understanding [33][45]. (4). By combining human factors, such as 'human's visual scope', with linguistic features, such as a keyword "wrench" in human NL instructions, empirical association model became environmental-context-sensitive, making a robot to understand a human NL instructions such as " deliver him a wrench" from the human perspective "human desired wrench is actually the human-visible wrench" [29][46][47]. The advantage of using association models in NLC is that the robot cognition level is improved by means of mutual knowledge compensation. With this association model, a robot can explore unfamiliar environments by exploiting its existing knowledge.

(ii). Interpreted models

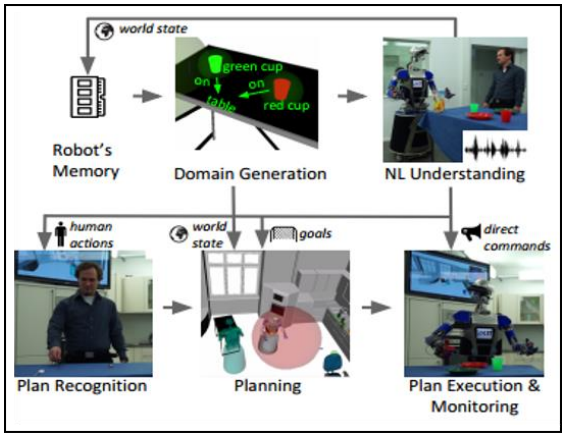

Fig.5. An typical interpreted model for NL instruction understanding [48]. Robot memory, real-world states and human NL instructions were integrated to instruct a robotic with plan executions.

Human requests are usually situated, which means human NL expressions are with default environmental preconditions, such as 'cup is dirty, a driller is missing, robot is far from a human'. Human NL instructions are closely correlated with situation-related information, such as human tactile indication (tactile modality), human hand/body pose (vision modality and motion dynamics modality), and environmental conditions (environment sensor modality).

To accurately understand human NL instructions, *interpreted models* are developed to integrate information from multi-modalities, instead of merely from NL modality. The rationale of interpreted models is that a human is considered to be dependent with their surrounding environment and better understanding of human needs to be environmentally context aware. With multi-modality models, information from different modalities related to human, robot, and their surrounding environment was aligned to establish semantic corrections [53][54]. Using NL instructions and human-related features to understand human NL instructions, typical features beyond linguistic features considered in single-modality models also include follows: (1) individual identity detected by RFID sensors [50]; (2). touch events detected by tactile sensors [51]; (3). facial expressions (joy, sad) [55], and hand poses detected by computer vision systems [56]; (4). human head orientations detected by motion tracking systems [57]. Supported by rich information from multi-modality information, typical problems tackled for NLC include complex-instruction understanding [52], human-like cooperation [57], human social behavior understanding, and mimicking [45]. For multi-modality models using environment and robot-related features to understand human NL instructions in NLC, typical features also include the following: (1). spatial object-robot relations indicated by human hand directions [52]; (2). temporal robot-speech-and-head-orientation dependencies measured by computer vision systems [57]; (3). object visual cues detected by cameras [58][59]; (4). robot sensorimotor behaviors monitored by both motion systems and computer vision systems [48]. Supported by rich information from these features, typical problems tackled in NLC include real-time communication, context-sensitive cooperation (sensor-speech alignment), machine-executable task plan generation, and implicit human request interpretation. Typical algorithms used for constructing multi-modality models include hidden Markov model (HMM) for modeling hidden probabilistic relations among interpreted linguistic features [58][60], Bayesian Network for modeling probabilistic transitions among task-execution steps [61][62], and first-order logic for modeling semantic constraints among interpreted linguistic features [49][63]. These algorithms integrate different modalities with appropriate contribution distributions and extract contributive feature patterns among modalities. Multi-modality models have three potential advantages in understanding human NL instructions. (1) By exploring multi-modality-information sources, rich information can be extracted for an accurate NL instruction understanding. (2) Information in one modality can be compensated by information learned from other modalities for better NL disambiguation. (3) Consistency of multi-modality information enables mutual confirmations among knowledge from multiple modalities. A reliable NL command understanding could be conducted. Supported by these advantages, multi-modality models have the potential to understand complex plans and various users, and to perform practical NL instruction understanding in real-world NLC situations.

B. *Model comparison*

Literal models, which use basic linguistic features directly from human NL instructions, are shallow literature-level understanding. While interpreted models, which use multi-modality features interpreted from human NL instructions, is a comprehensive connotation-level understanding. Each of them has unique advantages, therefore suitable for different application scenarios. For literal models, they are good at scenarios with simple procedures and clear work assignments, such as robot arm control and robot pose control. For interpreted models, they are good at scenarios with involvements of daily commonsense, human cognitive logics, rich domain information, such as object physical property assisted object searching, intuitive machine-executable plan generation, as well as vision-verbal-motion-supported object delivery. From literal models to interpreted models, robots have been more closely integrated with humans both physically and mentally. This integration enables a robot to accurately understand both human requests and practical environments, improving the effectiveness and naturalness of NLC.

TABLE. I. SUMMARY OF NL INSTRUCTION UNDERSTANDING METHODS

| | Literal models | | Interpreted models |
|---|---|---|---|
| | **Grammar** | **Association** | |
| **Knowledge format** | linguistic structures | meaningful concepts | semantic correlations |
| **Algorithms** | first-order logic | ontology tree | typical classification algorithms (NB, SVM), first-order logic |
| **User adaptability** | low | low | high |
| **Tackled problems** | initially understand logic relations, temporal and spatial relations in execution processes | specify abstract executions into machine-executable executions | complex task instruction understanding, human-like human-robot cooperation, context-sensitive cooperation |
| **Advantages** | performance is good and steady in trained situations | model human cognitive process, scaling up robot knowledge | rich cooperation-related information is involved. information is more reliable. |
| **Disadvantages** | exhaustive listing of NL instructions, time-consuming and labor-intensive | lacking standards for concept interpretation and interpretation evaluation | difficult to combine different-modality features, difficult to extract important NL features |
| **Typical references** | [30][31][37][38][39] | [40][41][43][44][45] | [58][60][61][62][63] |

*C. Open problems*

Although robots using grammar models have an initial capability of understanding human NL instructions during cooperation, the drawback is that feature correlations needed for understanding have been exhaustively listed. It is difficult to summarize all the likely-encountered grammar rules. Compared with grammar models, association models give more cooperation-related knowledge to a robot by exploiting associations among literal features. Even though the association model could interpret abstract linguistic features into detailed execution plan, it still suffers from incorrect association problems. These open problems are decreasing NL instruction understanding accuracy and further decreasing robot adaptability.

Although interpreted models are capable of comprehensively understanding human NL instructions by considering practical environment conditions, it is difficult to combine different types of modalities such as motion, speech, and visual cues with an appropriate manner to reveal practical contribution distributions for different modalities. Second, it is difficult to extract contributive features for describing both distinctive and common aspects of one modality in understanding NL instructions. Third, the overfitting problem still exists when using multi-modality information to understand NL instructions. NL instruction understanding based on different modalities could be mutually conflicting, thereby preventing the practical implementation of multi-modality models. Model details are shown in Table I.

IV. NL-BASED EXECUTION PLAN GENERATION

With task knowledge extracted in NL instruction understanding, it is critical to use the task knowledge to plan robot executions in NLC. Models for NL-based execution plan generation ('generation model' for short) are developed for formulating robot execution plans, theoretically supporting robots to cooperate with humans in appropriate manners. In these models, previously-learned piecemeal knowledge are organized with different algorithm structures. Different algorithms enable the models with different cooperation manners under various human-robot cooperation scenarios. For example, dynamic models supported by HMM enable real-time NL understanding and execution, while static models supported by Naïve Bayesian enable spatial human-robot relation exploration. During a plan generation, correlations among NLC-related knowledge, such as execution steps, step transitions, and actions, tools, or locations—as well as their temporal, spatial, and logic relations are defined. Regarding reasoning mechanisms, generation models have three main types: probabilistic models, logic models, and cognitive models.

*A. Models*

To enable robots with cooperation associative capability, in which a likely plan is inferred, and appropriate tools and actions are recommended, *probabilistic models* were developed based on probabilistic dependencies, shown in Fig. 6. To enable robots with logical reasoning capability, in which internal logics among execution procedures are followed, logic models were developed based on ontology and first-order logics, shown in Fig. 7. To enable robots with cognitive thinking capability, in which plans are intuitively made and adjusted, cognitive models were developed based on weighted logics, shown in Fig. 8.

(i). Probabilistic model

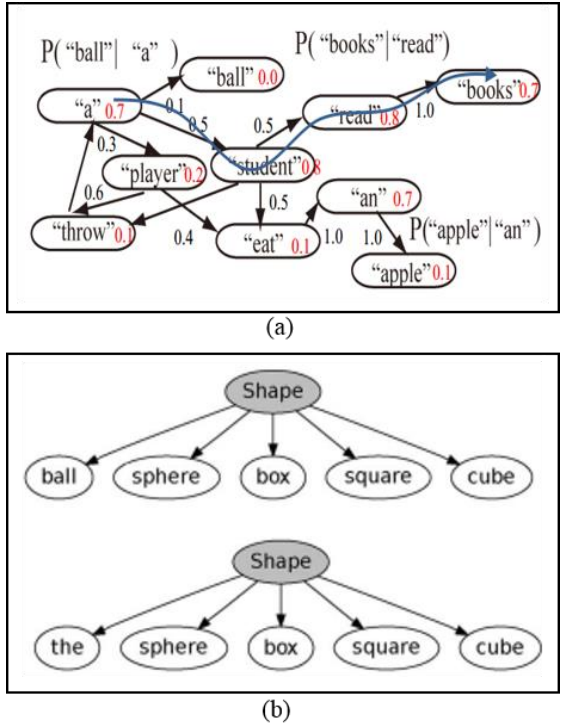

Fig. 6. Typical probabilistic models. (a) [64] is a HMM model, in which NLC task's potential execution sequences are modeled by hidden Markov statuses. (b) [65] is a naïve Bayesian model, in which observations "object size, object shape" and their conditional correlations such as "size-big, shape-ball, …" are combined to form joint-probability correlations such as "object-size-shape, ….".

To enable a robot with cooperation planning based on various observations, *joint-probabilistic BN methods* are developed. By using a single joint probability $p(x, y)$, a robot could use the probabilistic association $p$ between a human NL instruction $y$, such as "move", and one execution parameter $x$, such as object "ball", to plan simple cooperation such as object placement 'move ball' [65]. Typical joint-probability associations in NLC include activity-object associations, such as "drink-cup" [29], activity-environment associations such as "drink - hot day" [66], and action-sensor associations [67]. During the generation of cooperation strategies, a single joint-probabilistic BN association is used as independent evidence to describe one semantic aspect of a task. For using multiple joint-probabilistic associations $\prod_i p(x_i, y_i)$, interpreted linguistic features of NLC task are collected from various NL descriptions and sensor data, describing relatively complex plans. Typical methods using multiple joint-probability associations include Viterbi algorithm [67], Naïve Bayesian (NB) algorithm [66] and Markov Random Field (MRF) [68]. With these algorithms, the most complete plan described in human NL instructions are selected as a human-desired plan. With multi-joint-probabilistic BN models, tackled problems are as follows. (1). Modeling plans by extracting linguistic features, such as NL instruction patterns [68][69]; enriching cooperation details by aligning multiple types of sensor data, such as speech meaning, task execution statuses, and robot or human motion status [36]; making flexible plans by specifying verbally-described tasks with appropriate execution details, such as execution actions and effects [70][71]; intuitively cooperating with a human by integrating current NL descriptions with previous execution experiences [72]; accurate tool searching by associating theoretical knowledge, such as tool identities with practical real-world evidences, such as tools' colors and placement locations [73]. One common characteristic of probabilistic models, such as naïve Bayesian (NB, is that dependencies among task features are simplified to be fully or partially independent [66]. In practical situations, when a set of observations are made, evidences, such as speech, object, context, and action involved in cooperation, are usually not mutually-independent [74]. As for task plan representation, this simplification brings both negative effects, such as undermining the plan representation accuracy, and positive effects, such as preventing overfitting problems in plan-representation process. The common problem of multi-joint-probabilistic BN models is that temporal associations are ignored, limiting the implementations of real-time NLC.

To enable a temporal knowledge association for real-time cooperation planning, *Dynamic Bayesian Network* (DBN) was developed. With DBN, temporal dependencies $p(x_t|x_{t-1})$ propagated among NLC-related requests $x_t$ and object usages $x_{t-1}$ [75]. Given that the final format of DBN is the joint probabilistic form $p(y, x_t, x_{t-1}, ..)$, DBN is still a joint-probabilistic model. A widely-used DBN algorithm in NLC is hidden Markov mode (HMM) algorithm [76], which uses a Markov chain assumption to explore the hidden influence of previous task-related features on the current NLC status. The rationale of HMM in NLC is that human-desired executions, such as going to a position, grasping a tool, and lifting a robot hand, are decided by the previous cooperation ($x_{t-1}$), such as action sequence, and current cooperation $x_t$. These statuses include environmental conditions, task execution progress, and human NL instructions, as well as working statuses for the human and robot. HMM uses both observation probabilities (absolute probability p(x)) and transitions abilities (conditional probability p(Y/X)) for modeling associations P(x, y) among NLC-related knowledge [76][64]. With HMM models, tackled problems mainly include real-time task assignments [77], dynamic human-centered cooperation adjustment [64][78], accurate tool delivering by simultaneously fusing multi-view data such as NL instruction, shoulder coordinates, shoulders-elbows' 3D angle data, and hand poses [76][79]. Limited by Markov assumptions, HMM is only capable of modeling shallow-level hidden correlations among NLC-related knowledge. Moreover, given that hidden statuses need to be explored for HMM modeling, a large amount of training data is needed, limiting HMM implementations in unstructured scenarios with limited training data availability.

(ii). Logic model

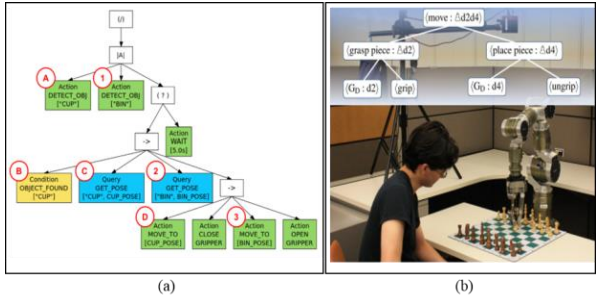

Fig.7. Typical logic models. In (a) [12], NLC tasks such as "finding a cup" is decomposed into different first-order logic constraints such as "cupAction + graspingPose → detectCup success". In (b) [35], hard logic relations such as "move=(grasp, place), …" is defined to control robots' motion in NLC.

To support a robot with rational logical reasoning of cooperation strategies, rather than merely conducting exhausting probabilistic inferences from various NL-indicated evidences, *logic models* were developed. Logic models teach robots using unviolated logic formulas to describe complex execution procedures which includes multiple actions and statuses. Unviolated logics usually are first-order logic formulas, such as "in possible worlds a kitchen is a region ($\forall w \forall x$(kitchen(w,x) → region(w,x)))" [80]. The rationale of logic models in NLC is that an NLC task is decomposed into sequential logic formulas by satisfying which specific NLC task could be accomplished. In a logic model, logics are equally important without contribution differences towards execution success. Logic relations, including tool usages, action sequences, and locations, are defined in the structure. Typical tackled problems include follows. (1). Autonomous robot navigation by using logic navigation sequences, such as going to a location "hallway" then going to a new location "rest room" [61][63]. (2). Environment uncertainty modeling by summarizing potential executions, such as "ground atoms (boolean random variables) eats(Dominik, Cereals), uses(Dominik, Bowl), eats(Michael, Cereals) and uses(Michael, Bowl)" [81]. (3). Robot action control by defining action-usage logics such as "move (grasp piece(location, grip), place piece(location, ungrip))" [61][82][83][84]. (4). Autonomous failure analysis by looking up first-order logic representations to detect the missing knowledge, such as "tool brush, action: sweep" [4][84]. (5). NL-based robot programming by using the grammar language, such as point(object, arm-side), lookat(object), and rotate(rot-dir, arm-side) [49]. The drawback of logic models in modeling NLC tasks is that logic relations defined in the model are hard constraints. If one logic formula was violated in practical execution processes, the whole logic structure would be inapplicable and the task execution would fail. This drawback limits models' implementation scopes and reduces a robot's environment adaptability. Moreover, hard constraints were defined indifferently, ignoring the relative importance of executions. The execution flexibility is undermined due to critical executions not being focused and trivial executions not being ignored when the NLC plan modifications are necessary.

(iii). Cognitive model

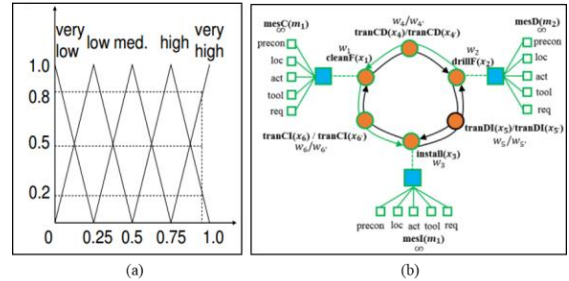

Fig.8. Typical cognitive models. In (a) **[87]**, human motion in NL instructions have been defined as fuzzy statuses "very low happy, very high happy, …" with different degree, reflecting human subjective attitudes in cooperation. In (b) [28], human's cognitive process in decision making is simulated by execution logics with different influence weights, based on which important logics with larger weights could be emphasized and trivial logics with smaller weights could be ignored. With this soft logic manner, the flexible cooperation between a human and a robot could be conducted.

Neural science research [85] and psychology research [86] proved that a cognitive human planning is not a sensorimotor transforming, instead a goal-based cognitive thinking. This reasoning is reflected on that cognitive thinking of cooperation is not relying on specific objects and specific executions, instead it is merely relying on goal realization. Based on this theory, another generation model category is summarized as a cognitive model. Human-like robot cognitive planning in NLC is reflected in flexibly changing execution plans (different procedures), adjusting execution orders (same procedures, different orders), removing some less-important execution steps (same procedures, less steps), and adding more critical executions procedures (similar procedures, similar orders).

To develop human-like robot cognitive planning for robust NLC, *cognitive models* are developed by using soft logic, which is defined by both logic formulas and their weighted importance. A typical cognitive model is Markov logic network (MLN) model. MLN represents NLC task in a way such as "$0.3\text{Drill}(mes_1)$ ∧ $0.3\text{TransitionFeasible}(mes_1, mes_2)$ ∧ $0.3\text{Clean}(mes_2)$ ⇒ $0.9\text{Task}(mes_1, mes_2)$" [4], imitating the human cognition process in task planning. In this model, single execution steps and step transitions were defined by logic formulas, which could be grounded into different logic formulas by substituting real-world conditions. With this cognitive model, a flexible execution plan can be generated by omitting non-contributing and weak-contributing logic formulas, and involving strong-contributing logic formulas. Different from hard constraints in logic models, constraints (logic formula) in MLN are soft. These soft constraints mean when human NL instructions are partially obeyed by a robot, the task could still be successfully executed. Typical tackled problems include using MLN to generate a flexible machine-executable plan from human NL instructions for autonomous industrial task execution [28], NL-based cooperation in uncertain environments by using MLN to meet constraints from both robots' knowledge availability (human-NL-instructed knowledge) and real-world's knowledge requirements (practical situation conditions) [81][88]. The advantage of using cognitive models in NLC is that soft logic is relatively like a human's cognitive process reflected in human NL instructions during cooperation. It helps a robot with intuitive cooperation in unfamiliar situations by modifying, or replacing, and executing plan details, such as tool or action usages, improving robots' cognition levels and enhancing its environment adaptability. The major drawback is that MLN is still different from human cognitive processes to consider logic conditions at a deep level to enable plan modification, new plan making, and failure analysis. Logic parameters for analyzing real-world conditions are still insufficient to imitate logic relations in the human mind, thereby limiting robots' performances in adapting to users and environments.

*B. Model comparison*

Usually the probabilistic model is conducted in an end-to-end manner, which directly reasons cooperation strategies from observations, ignoring internal correlations among execution procedures. A logic model uses a step-by-step manner, with which ontology correlations and temporal or spatial correlations among execution procedures are explored, enabling process reasoning for intuitive planning. The cognitive model also uses a step-by-step model. Including logic correlations, the cognitive model also explores relative influences of execution procedures, enabling a flexibly plan adjustment. For the probabilistic model, it is good for scenarios with rich evidence and single objective goal, such as tool delivery and navigation path selection. For the logic model, it is good for scenarios with either poor evidence or multiple objective goals, such as assembly planning and cup grasping planning. For the cognitive model, it is good for rich or poor evidence and multiple subjective goals, such as human emotion guided social interaction, and human preference based object assembly.

*C. Open problems*

Probabilistic models lack explorations of indirect human cognitive processes in NLC, limiting naturalness of robotic executions. Logic models is inflexible and incapable of simulating a human's intuitive planning in real-world environments. The cognitive model is close to a human's cognitive process in simulating flexible decision-making processes. However, cognitive models are still suffering from two types of shortcomings. One shortcoming is that cognitive process simulation is still not a cognitive process because the fundamental theory of cognitive process modeling is lacking insufficient support for a human-like task execution [81]. The second problem is the difficulties of cognitive model learning. Different individuals have different cognitive processes, thus making it difficult to learn a general reasoning model. Model details are shown in Table II.

V. KNOWLEDGE-WORLD MAPPING

With understanding of NL language and execution plans, it is critical for a robot to use this knowledge in practical cooperation scenarios. Knowledge-world mapping methods are developed to enable intuitive human-robot cooperation in real-world situations. The general process of knowledge-world mapping is shown in Fig. 12. Considering the different implementation problems, knowledge-world mapping methods include two main types: theoretical knowledge grounding and knowledge gap filling. Theoretical knowledge grounding methods accurately mapped learned knowledge items, such as objects, spatial/temporal logic relations, into corresponding objects and relations in real-world scenarios. Gap filling methods detect and recommend both the missing knowledge, which is needed in real-world situations, but has not been covered by theoretical execution plans, as well as real-world- inconsistent knowledge, which is provided by a human, but could not find corresponding things in practical real-world scenarios.

TABLE. II. METHOD SUMMARY OF NL-BASED EXECUTION PLAN GENERATION

|  | Probabilistic Model | | Logic Model | Cognitive Model |
|---|---|---|---|---|
|  | joint-probabilistic BN | Dynamic Bayesian Network | | |
| **Knowledge format** | joint probabilistic correlations | conditional probabilistic correlations | logic formulas | logic formulas, their weighted influences |
| **Algorithms** | joint BN, NB, MRF | conditional BN, Viterbi Algorithm, HMM | First-order Logic, Ontology Tree | MLN, Fuzzy Logic |
| **User adaptability** | moderate | moderate | low | high |
| **Tackled problems** | modeling meaning distributions on NL instructions, aligning multi-view sensor data, action and tool recommendation | meaning disambiguation, entity-sensor data mapping, human-attended object identification, real-time uncertainty assessment | autonomous robot navigation, environment uncertainty modeling, autonomous execution failure diagnosis, NL-based robot programming | support a flexible machine-executable plan implementation, task execution in unknown environments |
| **Advantages** | good at representing a complete plan | good at distinguishing plans | strong logic correlations among execution steps | flexible task plan, human cognitive process imitating, strong environment/user adaptability |
| **Disadvantages** | weak capability in modeling the mutual distinctiveness among tasks. | weak capability in representing a complete task; rely on large amount of training data | inflexible task execution, weak environment adaptation | parameters are difficult to learn, the current soft logic is still far from a human cognitive process |
| **Typical references** | [65][67][69] [70][71] | [75][76][77] [78][79] | [80][61][63] [82][83] | [28][81][88] |

*A. Models*

(i). Theoretical knowledge grounding

To accurately map theoretical knowledge to practical things, *knowledge grounding methods* are developed. In these methods, a knowledge item is defined by properties, such as visual properties "object color and shape" captured by RGB cameras, motion properties "action speed" captured by motion tracking systems and execution properties "tool usage and location" captured by RFID. Different from the direct symbol mapping method, which has an element-mapping manner, the general property mapping method has a structural mapping manner. The rationale of these methods is that a knowledge item can be successfully grounded into the real world by mapping its properties. The properties were collected by using methods such as 'semantic similarity measurement' [90], which can establish correlations between an object and their corresponding properties. One typical mapping method by using general property mapping is semantic map [91][92]. Theoretical indoor entities such as rooms and objects are identified by meaningful real-world properties, such as location, color, point cloud, spatial relation "parallel", neighbor entities, constructing a semantic map with both objective locations and semantic interpretations 'wall, ceiling, wall, floor'. For detecting visual properties in the real world, RGBD cameras are usually used. For spatial relations, it is detected by laser sensors, and motion tracking systems. By identifying these properties in the real-world, an indoor entity is identified, enabling an accurate robotic navigation in real-world NLC. Other typical mapping method also include: object searching by using NL instructions (detected by microphones) as well as visual properties such as object color, size and shape (detected by motion tracking systems and cameras) [93]; executing NL-instructed motion plan such as "pick up the tire pallet" by focusing on realizing actions "drive, insert, raise, drive, set" [94]; identifying human-desired cooperation places such as "lounge, lab, conference room" by checking spatial-semantic distributions of landmarks such as "hallway, gym, …" [95]. With mapping methods, knowledge could be mapped into real world in a flexible manner, in which only parts of properties need to be mapped for grounding a theoretical item into a real-world thing. This manner could improve a robot's adaptability towards users and environments. The limitation is that these mapping methods still use predefinitions to give a robot knowledge, reducing the intuitiveness of human-robot cooperation.

(ii). Knowledge gap filling

A theoretical execution plan defines an ideal real-world situation. Given unpredicted aspects in a practical situation, even if all defined knowledge has been accurately mapped into the real-world, it is still challenging to ensure the success of NLC by providing all knowledge needed in a practical situation. Especially in real-world situations, human users and environment conditions vary, causing the occurrences of knowledge gaps, which are knowledge required by real-world situations but are missing from a robot's knowledge database.

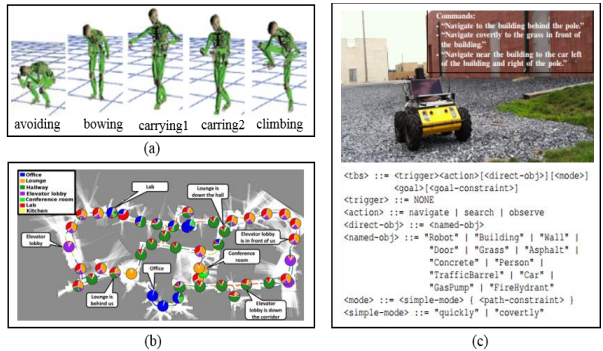

Fig.12. Typical methods for theoretical knowledge grounding. In (a) [89], predefined motion behaviors such as "avoiding, bowing, carring, .." are directly associated with their corresponding symbolic words. In (b) [92], special features such as "kitchen location, lab locations, …" are considered to identify human-desired paths. In (c) [9], a navigation task is specifically interpreted by real-world conditions such as "action: observe. named-obj:robot/building, … mode:quickly. … ".

To ensure the success of a robot's executions, *knowledge gap filling methods* are developed to fill in these knowledge gaps. There are three main types of knowledge gaps: (1). Environment gaps, which are constraints such as tool availability and space or location limitations imposed by unfamiliar environments [96]; (2). Robot gaps, which are constraints such as a robot's physical structure strength, capable actions and operation precision [97]; (3). user gaps, which are missing information caused by abstract, ambiguous, or incomplete human NL instructions [98][99]. Filling up these knowledge gaps enhances robot capability in adapting dynamic environments and various tasks or users. Knowledge gap filling is challenging in that it is difficult to make a robot aware of its knowledge shortage in specific situations, and it is difficult to make a robot understand how missing knowledge should be compensated for successful task executions.

The first step of gap filling is gap detection. Gap detection methods mainly include the following: (1) hierarchical knowledge structure checking, which detects knowledge gaps by checking real-world-available knowledge from top-level goals to low-level NLC execution parameters defined in a hierarchical knowledge structure [38][97], (2) knowledge-applicability assessment, which detects knowledge gaps by checking the similarities between theoretical scenarios and real-world scenarios [48][97], and (3) performance-triggered knowledge gap estimation, which detects knowledge gaps by considering the final execution performances [99][100]. Hierarchical knowledge structure checking has the rationale that if desired knowledge defined in a knowledge structure is missing in real-world situations, then knowledge gaps exist. Knowledge applicability assessment has a rationale that if the NLC situation is not similar with the previously-trained situations, then knowledge gaps exist. Performance-triggered knowledge gap estimation has a rationale that if the final NLC performances of a robot is not acceptable, then knowledge gaps exist. In this detection stage, execution plan provides reasoning mechanisms. While real world provides practical things such as objects, locations, human identities, and relations such as spatial relations and temporal relations, which are detected by perceiving systems.

The second step of gap filling is gap filling. Gap filling methods mainly include: (1). using existing alternative knowledge such as "brush" in the robot knowledge base to replace inappropriate knowledge such as "vacuum cleaner" in NLC tasks such as "clean a surface" [4][100]; (2). using general commonsense knowledge "drilling action needs driller" in a robot database to satisfy the need for a specific type of knowledge such as "tool for drilling a hole in the install a screw task" [100][97]; (3). asking knowledge input from human users by proactively asking questions such as "where is the table leg" [99][101][102]; (4). autonomously learning from the internet for recognizing human daily intentions, such as 'drink water, wash dishware' [101][102]. In gap filling stage, execution plan describes the needed knowledge items. Real world provides practical objects as well as robot performance monitoring.

*B. Model comparison*

Knowledge grounding model and knowledge gap filling model are two critical steps for a successful mapping between NL-instructed theoretical knowledge and real-world cooperation situations. For knowledge grounding models, the objective is strictly mapping NL-instructed objects and logic relations into real-world conditions. It is a necessary step for all the NLC application scenarios, such as human-like action learning, indoor and outdoor cooperative navigation. For knowledge gap filling models, the objective is to detect and repair missing or incorrect knowledge in human NL instructions. It is only necessary when human NL instruction cannot ensure successful NLC under given real-world conditions. Typical scenarios include: daily assistance such as serving drink, where information such as correct types of 'drink', 'vessel' and default places for drink delivery is missing; cooperative surface processing where execution procedures are incorrect and tools are missing.

TABLE. III. SUMMARY OF KNOWLEDGE-WORLD MAPPING METHODS

|  | Theoretical knowledge grounding | Knowledge gap filling | |
| --- | --- | --- | --- |
|  |  | Hierarchical Knowledge Structure Checking | Performance-Triggered Knowledge Gap Estimation |
| **Knowledge Format** | real-world objects, spatial/temporal/logic correlations | real-world objects, spatial/temporal/logic correlations | real-world objects, spatial/temporal/logic correlations |
| **Algorithms** | typical classification algorithms | first-order logic, ontology tree | typical classification algorithms |
| **User adaptability** | low | middle | middle |
| **Tackled problems** | indoor routine identifying, accurate object searching, scene understanding | detecting gaps among robots, users, and robots | detecting gaps among robots, users, and robots |
| **Advantages** | flexible knowledge usage, improving robots' environment adaptability. | improving smoothness of task executions | improving the success rate of task executions |
| **Disadvantages** | predefined manner limits robots' environment adaptability, execution naturalness and intuitiveness | difficult to decide which knowledge can miss | difficult to decide which gaps lead to the failure of task executions |
| **Typical References** | [94][92][93][95] | [72][100][97] | [97][101][102] |

*C. Open problems*

A typical problem of theoretical knowledge grounding is the non-executable-instruction problem. Human NL-instructed knowledge is usually ambiguous that NL-mentioned objects are too ambiguous to be identified in real world; abstract that high-level cooperation strategies are difficult to be interpreted into low-level execution details; information-incomplete that important cooperation information such as tool usages, action selections, and working locations are partially ignored; real-world inconsistent that human NL-instructed knowledge is not available in real world. These non-executable problems limit practical executions of human NL-instructed plans. One type of cause of non-executable-instruction problems include intrinsic NL characteristics, such as omitting, referring, and simplifying, as well as human speaking habits, such as different sentence organizations and phrase usages. Another type of cause is the lack of environment understanding. For example, if object-related information such as availability, location and distances to a robot or human was ignored, it is difficult for a robot to infer which object a human user needs [29].

For knowledge gap filling, when a robot queries knowledge from either a human or open knowledge databases such as openCYC [103], the scalability is limited. For a specific user or a specific open knowledge database, available contents are insufficient to satisfy general knowledge needs in various NLC executions. The time and labor cost is high, further limiting knowledge supports for NLC. Model details are shown in Table III.

VI. CONCLUSION & FUTURE WORK

This paper reviewed state-of-the-art methodologies for realizing natural-language-based human-robot cooperation (NLC). With in-depth analysis of application scenarios, method rationales, method formulations and current challenges, as well as research of using NL to push forward the limits of human-robot cooperation was summarized from a high-level perspective. This review paper mainly categorized a typical NLC process into three steps: NL instruction understanding, NL-based execution plan generation, and knowledge-world mapping. With these three steps, a robot can communicate with a human, reason about human NL instruction, and practically provide human-desired cooperation according to human NL instructions.

Future work of NLC research will improve accuracy of NL instruction understanding, flexibility of NL-based plan generation, and effectiveness of NL-based knowledge-world mapping. Therefore, to achieve these goals, potential NLC research could be: (1). exploring human cognitive process from NL instructions, (2). reducing robot knowledge cost by learning from web NL resources, and (3). personalizing robot by daily conversations, etc.

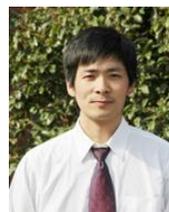
**Rui Liu** received the M.S. degree in active vibration control from Shanghai Jiao Tong University, Shanghai, China in 2013. Since January, 2014, he has been a Ph.D. student in Mechanical Engineering at Colorado School of Mines, Golden, CO, USA. His current research interests include robot knowledge (learn, interpret, implement), Metal Additive Manufacturing, NLP, machine learning and deep learning.

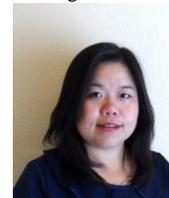
**Xiaoli Zhang** received the B.S. degree in Mechanical Engineering from Xi'an Jiaotong University, Xi'an, ShanXi, China in 2003, the M.S. degree in Mechatronics Engineering from Xi'an Jiaotong University in 2006, the Ph.D. degree in Biomedical Engineering from the University of Nebraska Lincoln, Lincoln, USA, in 2009.Since 2013, she has been an Assistant Professor in Colorado School of Mines. Her research interests include intelligent human-robot interaction, robotics system design and control, haptics.